# Stochastic Reinforcement Learning

Nikki Lijing Kuang
Department of Computer Science
and Engineering
University of California, San Diego
La Jolla, CA, USA
l1kuang@ucsd.edu

Clement H. C. Leung
Centre for Applied Informatics
Victoria University
Melbourne, Victoria, Australia
clement.leung@vu.edu.au

Vienne W. K. Sung
School of Continuing Education
Hong Kong Baptist University
Kowloon Tong, Hong Kong
wksung@hkbu.edu.hk

*Abstract*—In reinforcement learning episodes, the rewards and punishments are often non-deterministic, and there are invariably stochastic elements governing the underlying situation. Such stochastic elements are often numerous and cannot be known in advance, and they have a tendency to obscure the underlying rewards and punishments patterns. Indeed, if stochastic elements were absent, the same outcome would occur every time and the learning problems involved could be greatly simplified. In addition, in most practical situations, the cost of an observation to receive either a reward or punishment can be significant, and one would wish to arrive at the correct learning conclusion by incurring minimum cost. In this paper, we present a stochastic approach to reinforcement learning which explicitly models the variability present in the learning environment and the cost of observation. Criteria and rules for learning success are quantitatively analyzed, and probabilities of exceeding the observation cost bounds are also obtained.

*Keywords − reinforcement learning; multi-agent; non-deterministic environment; stopping rules; stochastic game; stochastic modeling; probability generating function*

## I. INTRODUCTION

In reinforcement learning (RL), an agent learns through the interaction with the dynamic environment to maximize its long-term rewards, in order to act optimally. Most of the time, when modeling real-world problems, the environment involved is non-stationary and noisy [1][4][6]. More precisely, the next state results from taking the same action in a specific state may not necessarily be the same but appears to be stochastic [2][7]. And the exploration strategies adopted in different categories of RL algorithms provide different levels of control to the exploration of unknown factors, which in turn give various possibilities to the learning results.

As a result, the observed rewards and punishments are often non-deterministic. For example, when one is trying out a new route to work, a shortening of the travel time may be regarded as a reward, while a lengthening of the same may be viewed as punishment. Likewise, when one is exploring a new advertising channel, a resultant significant increase in sales may be viewed as a reward, while failure to do so may be regarded as punishment. In situations like these, there are stochastic elements governing the underlying environment. In the new route to work example, whether one receives rewards or punishments depends on a variety of chance factors, such as weather condition, day of the week, and whether there happens to be road works or traffic accidents which may or may not be representative.

Such hidden variables are generally numerous and cannot be known or enumerated in a practical sense, and these tend to mask the underlying pattern. Indeed, if stochastic elements are absent, the learning problems involved could be greatly simplified and their presence has motivated early research in the area. As early as 1990s, mainstream research in RL, such as the influential survey assessing existing methods carried out by Kaelbling, *et al.* [2], and the Explicit Explore or Exploit ($E^3$) Algorithm to solve Markov Decision Process (MDP) in polynomial time [3], adopts the common assumption of a stationary environment within a RL framework. Later on, with further advances in RL, theoretical analyses addressing the concern of non-stationary environment attracted great interests. One of the works by Brafman and Tennenholtz introduces a model-based RL algorithm R-Max to deal with stochastic games [5]. Such stochastic elements can notably increase the complexity in multi-agent systems and multi-agent tasks, where agents learn to cooperate and compete simultaneously [6][10]. Autonomous agents are required to learn new behaviors online and predict the behaviors of other agents in multi-agent systems. As other agents adapt and actively adjust their policies, the best policy for each agent would evolve dynamically, giving rise to non-stationarity [8][9].

In most of the above situations, the cost of a trial or observation to receive either a reward or punishment can be significant, and preferably, one would like to arrive at the correct conclusion by incurring minimum cost. In the case of the advertising example, the cost of advertising can be considerable and one would therefore like to minimize it while acquiring the knowledge whether such advertising channel is effective. Similarly, in RL algorithms, we are always in the hope to rapidly converge to an optimal policy with least volumes of data, calculations, learning iterations, and minimal degree of complexity [11][12]. To do so, one should explicitly define the stopping rules for specifying the conditions under which learning should terminate and a conclusion drawn as to whether the learning has been successful or not based on the observations so far.

The problem of finding termination conditions, or stopping rules, is an intensive research topic in RL, which is closely linked to the problems of optimal policies and policy convergence [13]. Traditional RL algorithms mainly aim for relatively small-scale problems with finite states and actions. The stopping rules involved are well-defined for each category of algorithms, such as utilizing Bellman Equation in *Q*-learning [14]. To deal with continuous action spaces or state spaces, new algorithms, such as the Cacla algorithm [15] and CMA-ES algorithm [16], are developed with specific stopping rules. Still, most studies on stopping rules are algorithm-oriented and do not have a unified measurement for general comparison.

In this paper, we present an approach to RL which explicitly incorporates the stochastic learning environment. Section II presents the fundamental model of a predefined general stopping rule. The learning success based on the rewards ratio is then studied in Section III. Based on the stochastic model, Section IV analyzes the probability of exceeding cost bounds. Section V views the occurrences of positive and negative rewards from the perspective of competing multi-agents, and the final conclusions are drawn in Section VI.

## II. A Probabilistic Learning Framework With A Fixed Number of Rewards

Here, we are dealing with a sequence of learning observations, each of which either incorporates a positive reward or negative reward (i.e. punishment). We let $p$ and $q$, with $p + q = 1$, denote the probabilities of receiving a positive reward or negative reward respectively for a given observation. For example, if $p > q$, then clearly the final conclusion should be that the learning is successful, and in the case of our examples, the new route should be adopted as well as the new advertising channel should be used. An error often committed is that when the first few observations are all negative, one would terminate prematurely and conclude that the learning episode is a failure. Let us consider the stopping rule:

**Rule A**: *An agent stops a learning episode upon observing r positive rewards. The learning episode can be concluded as a success or a failure according to whether the number of positive rewards is sufficiently greater than the number of negative rewards.*

This is in fact a family of rules since there are two parameters that need to be specified. The first is $r$, and the second is related to "sufficiently greater", both of which need to be indicated in a more detailed and quantitative manner.

Consider the probabilistic model of *Rule A*. Let the random variable $T$ be the number of observations preceding the first positive reward; i.e. $T$ may be regarded as the waiting time to the first positive reward, measured in number of observations, then

$$\Pr[T = k] = pq^k, \quad k = 0, 1, 2, 3, \ldots \tag{1}$$

The probability generating function $G(z)$ of $T$ is given by

$$G(z) = \sum_{k=0}^{\infty} \Pr[T = k] z^k = p \sum_{k=0}^{\infty} q^k z^k = \frac{p}{(1-qz)}. \tag{2}$$

Note that after the occurrence of the first reward, the process probabilistically repeats itself again, so that we have for the waiting time $W_r$ of the $r$th positive reward

$$W_r = \sum_{k=1}^{r} T_k, \tag{3}$$

where each $T_k$ has the same distributional characteristics as $T$. From [17], the probability generating function of $G_r(z)$ corresponding to $W_r$ may be obtained

$$G_r(z) = G_1(z)^r = \left[\frac{p}{(1-qz)}\right]^r. \tag{4}$$

To get a better understanding of behavior specified above, it is useful to obtain the average waiting time $W_r$ and its variance when $r$ positive rewards are observed. From (4), the mean and variance of $W_r$ can be derived

$$E[W_r] = G'_r(1) = \frac{rq}{p}, \tag{5}$$

$$\text{Var}[W_r] = G''_r(1) + G'_r(1) - G'_r(1)^2 = \frac{rq}{p^2}. \tag{6}$$

Furthermore, the probabilities $\Pr[W_r = k]$ may be readily obtained from the expansion of (4) so as to study the probabilities for various waiting time,

$$\Pr[W_r = k] = \binom{-r}{k} p^r (-q)^k, \quad k = 0, 1, 2, 3, \ldots \tag{7}$$

Since $W_r$ is the sum of independent identically distributed random variables, when $r$ is appreciable, it may be approximated by the normal variate by virtue of the Central Limit Theorem [17], so that

$$W_r \sim N\left(\frac{rq}{p}, \frac{rq}{p^2}\right), \tag{8}$$

where $N(\mu, \sigma^2)$ denotes the normal distribution with mean $\mu$ and variance $\sigma^2$. Thus, the probability $\Pr[W_r > b]$ may be approximated by

$$\Pr[W_r > b] = \int_{\frac{bp-rq}{\sqrt{rq}}}^{\infty} \frac{1}{\sqrt{2\pi}} e^{-\frac{t^2}{2}}$$

$$= 1 - \Phi\left(\frac{bp-rq}{\sqrt{rq}}\right), \tag{9}$$

where $\Phi$ is the standard normal distribution.

## III. LEARNING SUCCESS BASED ON THE REWARDS RATIO

Let ratio $\rho$ to be the ratio of the average number of negative rewards to the number of positive rewards, which can be obtained from (5)

$$\rho(p) = \frac{E[W_r]}{r} = \frac{1-p}{p}. \quad (10)$$

It is interesting to note that the rewards ratio is linearly correlated to the probability ratio of negative rewards to positive rewards $q/p$. For successful learning, a necessary condition for an optimal policy is to have the number of negative rewards strictly less than the number of positive rewards or $\rho < 1$. From (10), this is equivalent to $q < p$, implying $p > 1/2$.

Generally, we wish to have a small value for $\rho$, so that the average number of negative rewards received is much less than the number of positive rewards, i.e. $q << p$. In Fig. 1, the change of $\rho(p)$ is shown as $p$ varies over the range (0,1), where $\rho$ reaches unity at the point (0.5, 1). We see that as $p$ tends to 0, the rewards ratio $\rho$ tends to infinity. It intuitively suggests that if the probability of receiving a single positive reward is small, the number of negative rewards on average will be high. In order to have a small rewards ratio for successful learning, so as to maximize the relative number of positive rewards, $p$ should be significantly greater than 1/2.

As mentioned above, the indication of "sufficiently greater" in *Rule A* requires quantification. Here, we pre-specify a particular fraction of negative rewards $\rho^*$ (< 1), so that upon the completion of a learning episode, i.e. $r$ positive rewards are observed, if the actual number of negative rewards $W$ is such that $W/r < \rho^*$, the single learning is concluded as successful; otherwise, it is regarded as a failure.

Possible values of $\rho^*$ may be 30%, 50% or 70%, depending how stringent is the learning correctness criterion. In any case, either type of error – i.e. wrongly concluding success or wrongly concluding failure – is unavoidable, as in standard statistical testing. In the case of $\rho^* = 100\%$, we have $W = r$, and it would be difficult to draw a definite conclusion concerning the learning outcome, and to be conservative, one would therefore conclude that the learning has been unsuccessful.

Looking at things from another angle, it is also possible to estimate $p$ from actual rewards ratio $W/r$ to and to draw a conclusion of the learning episode. From (10), solving for $p$, we have

$$\hat{p} = \frac{1}{1 + W/r},$$

and we see that $p > 1/2$ if $W/r < 1$. A more systematic approach to the estimation of $p$ using confidence intervals will not be examined here, and will be the subject of a separate study.

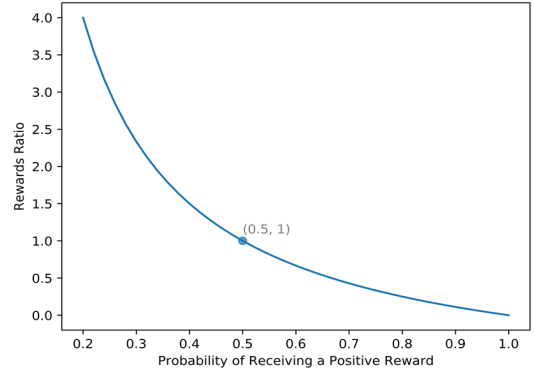

Figure 1. Changes of Rewards Ratio

## IV. PROBABILITY OF EXCEEDING COST BOUNDS

The mean value, being a single statistic, is often not sufficient as it fails to fully reflect any statistical fluctuations. In many cases, as in the advertising example, the cost of observation is significant. Let $c$ be the numerical representation of cost associated with an observation. Having specified $r$, a minimum observation cost of $rc$ must therefore be incurred. What is uncertain is the number of negative rewards obtained, and ideally to achieve minimum cost, this number should be bounded. Supposing we would only sustain a maximum cost of $bc$ for observing $b$ negative rewards, we shall estimate the probability $P_b$ that the learning cost for this component exceeding this bound. From (7) above, this is given by

$$P_b = 1 - \sum_{k=0}^{b} \Pr[W_r = k] = 1 - \sum_{k=0}^{b} \binom{-r}{k} p^r (-q)^k. \quad (11)$$

The computation associated with (11) is somewhat laborious. As indicated above, when the value of $r$ is large, we can make use of the normal approximation of (9). In many RL learning episodes, $r$ tends to be under 100, as a lengthy iteration time is not feasible and most learning algorithms aim to converge in minimum time.

Clearly, the selection of the maximum cost weight $b$ will have a significant impact on $P_b$. Very often, it is more meaningful to relate $b$ to $E[W_r]$ either additively or multiplicatively. Table I tabulates the values of $P_b$ for different values of $b$. The first part of Table I considers $b$ by adding a fixed value $d$, with $d = 5$ and $d = 10$, while the second part considers $b$ by multiplying by a fixed multiple $\alpha$, with $\alpha = 1.2$ and $\alpha = 1.5$; here, $b$ is rounded to the nearest integer. In the first part of Table I, we see that for either value of $r$, when $p$ is appreciably greater than $q$, the probability of exceeding cost bounds tends to be acceptably small, and this is especially so for $r = 20$. The reason is that, since $d$ is a fixed value, its relative contribution to $b$ increases as $p$ increases, produces a relatively large cost bound weight compared to the average one, and accordingly lowers the probability of exceeding the bound. However, in the second part of Table I, the difference between $E[W_r]$ and b decreases as E[$W_r$] decreases, so that $P_b$ tends to be large for higher values of $p$.

TABLE I. ANALYSIS OF PROBABILITIES OF EXCEEDING COST BOUNDS

| b Formula | r | p | q | E[$W_r$] | b | $P_b$ | $P_b'$ | Err |
|---|---|---|---|---|---|---|---|---|
| b = E[$W_r$] + d (d = 5) | 20 | 0.5 | 0.5 | 20.00 | 25 | 0.215 | 0.186 | 0.029 |
| | | 0.8 | 0.2 | 5.00 | 10 | 0.023 | 0.026 | 0.003 |
| | | 0.9 | 0.1 | 2.22 | 7 | 0.001 | 0.004 | 0.003 |
| | 50 | 0.5 | 0.5 | 50.00 | 55 | 0.309 | 0.279 | 0.030 |
| | | 0.8 | 0.2 | 12.50 | 17 | 0.127 | 0.108 | 0.019 |
| | | 0.9 | 0.1 | 05.56 | 11 | 0.014 | 0.017 | 0.003 |
| b = E[$W_r$] + d (d = 10) | 20 | 0.5 | 0.5 | 20.00 | 30 | 0.057 | 0.059 | 0.002 |
| | | 0.8 | 0.2 | 5.00 | 15 | 0.000 | 0.001 | 0.001 |
| | | 0.9 | 0.1 | 2.22 | 12 | 0.000 | 0.000 | 0.000 |
| | 50 | 0.5 | 0.5 | 50.00 | 60 | 0.159 | 0.147 | 0.012 |
| | | 0.8 | 0.2 | 12.50 | 22 | 0.008 | 0.011 | 0.003 |
| | | 0.9 | 0.1 | 05.56 | 16 | 0.000 | 0.000 | 0.000 |
| b = αE[$W_r$] (α = 1.2) | 20 | 0.5 | 0.5 | 20.00 | 24 | 0.264 | 0.226 | 0.038 |
| | | 0.8 | 0.2 | 5.00 | 6 | 0.345 | 0.253 | 0.092 |
| | | 0.9 | 0.1 | 2.22 | 2 | 0.556 | 0.380 | 0.176 |
| | 50 | 0.5 | 0.5 | 50.00 | 50 | 0.159 | 0.147 | 0.012 |
| | | 0.8 | 0.2 | 12.50 | 15 | 0.264 | 0.215 | 0.049 |
| | | 0.9 | 0.1 | 05.56 | 7 | 0.280 | 0.207 | 0.073 |
| b = αE[$W_r$] (α = 1.5) | 20 | 0.5 | 0.5 | 20.00 | 30 | 0.057 | 0.059 | 0.002 |
| | | 0.8 | 0.2 | 5.00 | 7 | 0.212 | 0.156 | 0.056 |
| | | 0.9 | 0.1 | 2.22 | 3 | 0.310 | 0.193 | 0.117 |
| | 50 | 0.5 | 0.5 | 50.00 | 75 | 0.006 | 0.010 | 0.004 |
| | | 0.8 | 0.2 | 12.50 | 19 | 0.050 | 0.048 | 0.002 |
| | | 0.9 | 0.1 | 05.56 | 8 | 0.163 | 0.121 | 0.042 |

In Table I, column $P_b'$ gives the exact calculation using (11), while column $P_b$ employs the normal approximation using (9). The absolute error between the exact calculation and the normal approximation is given by column *Err*. We see that the normal approximation is quite acceptable in most cases with absolute error less than 0.1. Note that no matter whether having b additively or multiplicatively related to E[$W_r$], a higher value of d or α always gives smaller absolute error. We therefore suggest that the approximation should only be used when $r$, $d$ and $α$ are sufficiently large.

## V. A COMPETING MULTI-AGENT LEARNING FRAMEWORK

In *Rule A* above, the termination of a learning episode is triggered whenever a fixed number of positive rewards $r$ is obtained, irrespective of the number of negative awards accumulated in the process of doing so. Sometimes, however, this may not be desirable, especially when an inordinate number of negative rewards have been accumulated, in which case, termination should take place earlier along with the conclusion of learning failure. Therefore, one is comparing the number of positive rewards gathered against the number of negative rewards, and the learning is concluded as success or failure according to which of these achieve the majority.

More precisely, this may be viewed as a multi-agent game with two competing agents *A* and *B*, in which *A* is responsible for giving out the positive rewards, while *B*, the negative rewards. This framework is not unlike the game theoretic approach in statistical decision theory, where both the statistician and nature are regarded as players in the game of estimation, and also this may be regarded as a kind of stochastic game [5]. While we shall focus on the agents *A* and *B*, we note that there is a further agent, the learner, so that three agents exist in this situation. Here, when an observation results in a positive reward, then *A* would gain a score of one, while when an observation results in a negative reward, then *B* would gain a score of one. When either score first reaches a given threshold $h$, then this will trigger a termination and the learning episode is concluded as success or failure according to which agent attains the threshold score first. Therefore, we have the following stopping rule:

**Rule B**: *A learning episode stops upon either agent, A or B, first reaching the threshold of h rewards. The learning episode can be concluded as a success or a failure according to which agent attains the threshold first.*

Here, without loss of generality, we shall let $h = 2m+1$ be odd, where $m$ is an integer, and similar to Section II, we let $p$ and $q$, with $p + q = 1$, signify the probabilities of receiving a positive reward and negative reward respectively for a particular observation. In other words, for a given observation, agent *A* wins with probability $p$, while agent *B* wins with probability $q$. In order to attain $h$ for either agent, the number of observations $\Omega$ will fall within the range

$$2m + 1 \leq \Omega \leq 4m + 1.$$

If $f_k$ represents the probability that *A* wins at observation number $4m+1-k$, which occurs if and only if *A* scored $2m$ successes in the first $4m-k$ observations, and subsequently score a final success, then $f_k$ is given by

$$f_k = \binom{4m - k}{2m} p^{2m+1} q^{2m-k}.$$

The probability that *A* reaches the threshold first, irrespective of the observation number, is therefore given by

$$P_m = \sum_{k=0}^{2m} f_k = \sum_{k=0}^{2m} \binom{4m - k}{2m} p^{2m+1} q^{2m-k}.$$

That is, $P_m$ gives the probability that the learning is successful (i.e. agent *A* wins) according to *Rule B*.

Table II computes $P_m$ for different values of $p$, $q$, and $m$. We see that, as expected, when $p = q = 1/2$, $P_m = 1/2$, since neither *A* nor *B* has any advantage over its opponent. As $p$ increases, however, $P_m$ will increase, reaching almost certainty as $p$ increases beyond 0.8. If we regard $p$ as a measure of $A$'s winning ability per trial, then when $p \gg q$, most trials will be scored by $A$, so that winning the entire game (i.e. reaching $h$ first) is almost a certainty, and this is especially so for higher values of $h$. It is interesting to see that when $h$ or $m$ is sufficiently high (e.g. $m=10$), a moderate advantage for $A$ (e.g. $p = 0.6$) is enough to almost guarantee success. On the other hand, $1-P_m$ gives the probability that agent $B$ wins, where the measure of $B$'s winning probability per trial is given by $q$. For instance, when $q=0.4$, then $B$ stands a chance of around 27% of winning the game when $m=2$, and a chance of winning of around 10% when $m=10$.

TABLE II. PROBABILITIES OF LEARNING SUCCESS

| m | p | q | $P_m$ | m | p | q | $P_m$ |
|---|---|---|---|---|---|---|---|
| 1 | 0.5 | 0.5 | 0.5000 | 5 | 0.5 | 0.5 | 0.5000 |
|   | 0.6 | 0.4 | 0.6826 |   | 0.6 | 0.4 | 0.8256 |
|   | 0.7 | 0.3 | 0.8369 |   | 0.7 | 0.3 | 0.9736 |
|   | 0.8 | 0.2 | 0.9421 |   | 0.8 | 0.2 | 0.9990 |
|   | 0.9 | 0.1 | 0.9914 |   | 0.9 | 0.1 | 1.0000 |
| 2 | 0.5 | 0.5 | 0.5000 | 10 | 0.5 | 0.5 | 0.5000 |
|   | 0.6 | 0.4 | 0.7334 |   | 0.6 | 0.4 | 0.9035 |
|   | 0.7 | 0.3 | 0.9012 |   | 0.7 | 0.3 | 0.9964 |
|   | 0.8 | 0.2 | 0.9804 |   | 0.8 | 0.2 | 1.0000 |
|   | 0.9 | 0.1 | 0.9991 |   | 0.9 | 0.1 | 1.0000 |

## VI. SUMMARY AND CONCLUSION

Since learning environments are often noisy and seldom static nor deterministic, the use of stochastic models in reinforcement learning is therefore an unavoidable necessity. Indeed, if stochastic elements are absent, the same outcome will always occur, obviating the need for repeated observations. Here, we explicitly represent the occurrence of positive and negative rewards probabilistically, where each positive reward is assumed to be independent, identically distributed.

In this paper, we first consider a situation where the cumulative number of rewards is pre-specified and fixed, which constitute the criterion for stopping the learning process. By examining the positive to negative rewards ratio, a meaningful conclusion of either learning success or failure may be arrived at. In most practical situations, the cost of observation can be significant, and this has been incorporated into our model, with the probabilities of exceeding the observation cost bounds also derived.

We next consider a multi-agent framework where the handing out of positive and negative rewards are viewed as being performed by agents. Thus, the final learning outcome is determined by a kind of stochastic game with the agents competing against each other. The stopping criterion here is determined by when and how the game is won. The respective probabilities of learning success and failure are also explicitly derived. Closed-form expressions of other relevant measures of interest are obtained.

In this study, we have assumed that positive rewards and negative rewards occur independently. In future, it may be useful to relax this assumption and incorporate single-step or multi-step Markov dependency into the analysis.